# Neural Network for Low-Memory IoT Devices and MNIST Image Recognition Using Kernels Based on Logistic Map

**Andrei Velichko**

Institute of Physics and Technology, Petrozavodsk State University, 31 Lenina Str., 185910 Petrozavodsk, Russia; velichko@petrsu.ru; Tel.: +7-8142-63-5773

**Abstract:** This study presents a neural network which uses filters based on logistic mapping (LogNNet). LogNNet has a feedforward network structure, but possesses the properties of reservoir neural networks. The input weight matrix, set by a recurrent logistic mapping, forms the kernels that transform the input space to the higher-dimensional feature space. The most effective recognition of a handwritten digit from MNIST-10 occurs under chaotic behavior of the logistic map. The correlation of classification accuracy with the value of the Lyapunov exponent was obtained. An advantage of LogNNet implementation on IoT devices is the significant savings in memory used. At the same time, LogNNet has a simple algorithm and performance indicators comparable to those of the best resource-efficient algorithms available at the moment. The presented network architecture uses an array of weights with a total memory size from 1 to 29 kB and achieves a classification accuracy of 80.3–96.3%. Memory is saved due to the processor, which sequentially calculates the required weight coefficients during the network operation using the analytical equation of the logistic mapping. The proposed neural network can be used in implementations of artificial intelligence based on constrained devices with limited memory, which are integral blocks for creating ambient intelligence in modern IoT environments. From a research perspective, LogNNet can contribute to the understanding of the fundamental issues of the influence of chaos on the behavior of reservoir-type neural networks.

## 1. Introduction

In the age of neural networks and Internet of Things (IoT), the search for new neural network architectures capable of operating on devices with small amounts of memory (10s of kB of RAM) is becoming an urgent agenda [1–3]. The constrained devices possess significantly less processing power and memory than a regular smartphone or modern laptop, and usually do not have a user interface [4]. The constrained devices form the basis for ambient intelligence (AmI) in IoT environments [5], and can be divided into three categories based on code and memory sizes: class 0 (less than 100 KB Flash and less than 1 KB RAM), class 1 ( ≈100 KB Flash and ≈ 10 KB RAM) and class 2 ( ≈ 250 KB Flash and ≈ 50 KB RAM) [6]. Smart objects, which are built on constrained devices, possess limited system resources, and these limitations prohibit the application of security mechanisms common in the Internet environment. Complex cryptographic mechanisms require significant time and high energy resources, and in addition, the storage of a large number of keys for secure data transition is not possible on the constrained devices. The storage issue presents significant challenges for the integration of blockchain [7] and artificial intelligence (AI) [1,8] with IoT and calls for new approaches and research efforts to resolve this limitation. Intelligent IoT devices should be able to process the incoming information without sending it to the cloud. Smart objects equipped with their own AI capabilities spend significantly less time on the analysis of incoming data and development of a final solution, creating new possibilities, for example, in the development of AmI in the medical industry [9], predicting the behavior of mechanisms [10], local semantic processing of video data [11] and "smart" services in e-tourism [12]. This approach facilitates the development of the concepts of smart spaces and fog computing, when devices detect each other, for example, using wireless technologies [13]; redistribute computational tasks; and optimize the distribution of responses [14,15]. Therefore, the integration of AI and IoT creates new social, economic and technological benefits.

Neural networks create the foundation for AI. The well-known types of neural networks are feedforward neural networks, convolutional neural networks and recurrent neural networks [16]. In addition to neural networks, the popular methods for classification and efficient prediction include tree-based algorithms [17] and the k-nearest neighbors algorithm (kNN) [18]. The problem of implementing compact and efficient algorithms for the operation of neural networks on constrained devices is the main constraining factor in the development of the integration of AI and IoT. Training of neural networks is the process of optimal selection of the weight coefficients of neuron couplings and filter parameters that occupy a significant amount of memory. It highlights the importance of the development of the methods for reducing the memory consumed by a neural network.

The prominent tree-based algorithms include the GBDT algorithms [19] and Bonsai from Microsoft [17], which allows efficient use of memory and runs on constrained devices with 2–16 kB RAM. An algorithm called ProtoNN [20], based on the kNN method, operates on devices with 16 kB RAM. Currently, effective memory redistribution algorithms in convolutional neural networks (CNN) are available and do not exceed 2 kB of consumed RAM [21], and demonstrated impressive results of classification accuracy ≈ 99.15% on MNIST-10 database. However, the complexity of the algorithms leads to the large size of the program itself. Algorithms based on the recurrent network architecture, such as Spectral-RNN [22] and FastGRNN [18], achieve ≈ 98% classification accuracy on MNIST-10 with a model size of ≈ 6 kB.

An alternative approach to reduce the number of trained weights is based on physical reservoir computing (RC) [23]. RC uses complex physical dynamic systems (coupled oscillators [24–26], memristor crossbar arrays [27], opto-electronic feedback loop [28]), or recurrent neural networks (echo state networks (ESNs) [29] and liquid state machines (LSMs) [30]), as reservoirs with rich dynamics and powerful computing capabilities. The couplings in the reservoir are not trained, but are specified in a special way. The reservoir translates the input data into a higher dimensionality of space (kernel trick [31]), and the output neural network, after training, can classify the result more accurately. The search for simple algorithms for simulating complex reservoir dynamics is an important research task.

The current study presents a new architecture for a neural network, where complex dynamics are simulated by the application of logistic mapping to the multilayer weights of a feedforward network. Previously, logistic mapping was used in neural networks as an activation function [32,33] and as a model object [34,35]. The application of logistic mapping significantly reduces the amount of memory used by the network without losing functionality. The study applies to the recognition of handwritten digits from MNIST-10 database.

The rest of the paper is organized as follows. In the next section, the architecture of the LogNNet network, an algorithm for finding weights using logistic mapping and a method for assessing the accuracy of classification of handwritten digits from the MNIST-10 database are described. Section 3 contains the results of the dependences of the classification accuracy on the key parameters of the neural network and the assessment of the efficiency of memory use. Three algorithms for calculating the reservoir weights, whose code is presented in Appendix A, allow a LogNNet configuration that occupies from 1 to 29 kB of memory and achieves a classification accuracy of 80.3–96.3%. In Section 4, discussion of the results reviews the correlation between the classification efficiency of the network and the Lyapunov exponent, which characterizes the chaotic behavior of the logistic mapping.

LogNNet is compared with other well-known algorithms, and suggestions to improve functionality and to reduce the amount of memory occupied are given. In conclusion, the study results indicate that the proposed neural network can be used to implement artificial intelligence based on constrained devices with limited memory, which are integral blocks for the creation of AmI and modern IoT environments.

## 2. Data and Methods

The database of handwritten digits MNIST-10 (available on Yan LeCun's Internet page [36]) was used for the study. The database consists of 70,000 images of handwritten digits from "0" to "9". Each image is 28 × 28 pixels in size (see Figure 1a). The database is divided into two sets. The first set consists of 60,000 images intended for training the network, and the second set of 10,000 images is used to test the network and to calculate classification accuracy. The image is presented in grayscale with the intensity of each pixel in the range from 0 to 255. Before inputting the network, a two-dimensional image was converted into a one-dimensional array $Y[i]$, where $i$ = 1–784, using special transformation T-patterns. The main types of T-patterns used are presented in Figure 1b–d. In addition, T-patterns can be used for reverse transformation from a one-dimensional array to a two-dimensional array.

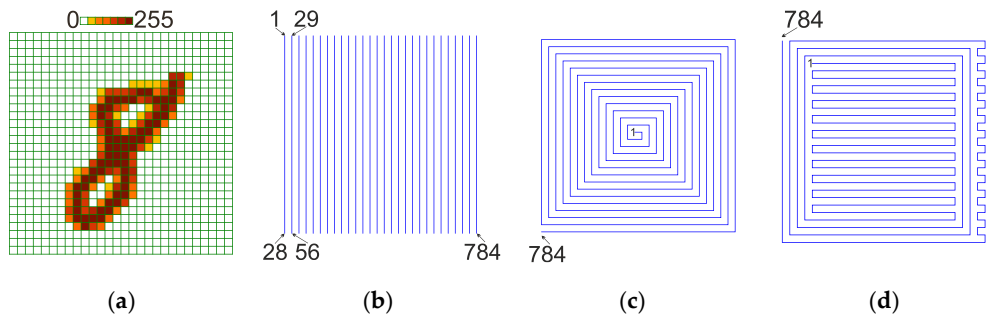

(a) (b) (c) (d)

**Figure 1.** An example of a 28 × 28 pixel image of the handwritten digit "8" from the MNIST-10 database (**a**). Patterns for transforming the image array: T-pattern-1 (**b**), T-pattern-2 (**c**), T-pattern-3 (**d**).

T-pattern-1 performs column-by-column input of the image; T-pattern-2 converts the image in a spiral (clockwise) form; and T-pattern-3 is a combination of line-by-line input of the central part of the image, with the addition of borders in a spiral (clockwise) form.

After the T-pattern transformation, the values of the array $Y$ are normalized by dividing each element of the array by 255, and a bias element $Y[0] = 1$ is added.

*Network Architecture*

The network architecture used in the current study is presented in Figure 2a. Data processing was performed similarly to the feedforward network (Figure 2b), according to the equations

$$S_h = f_h(Y \cdot W_1), \; S_{out} = f_{out}(S_h \cdot W_2), \qquad (1)$$

where $W_1$ and $W_2$ are the weight matrices; $S_h$ is the hidden layer with the number of active elements $P$ and the bias element $S_h[0] = 1$; $S_{out}$ is the output layer; $f_h$ is identity activation function, which is normalized in the range from −0.5 to 0.5; and $f_{out}$ is logistic activation function.

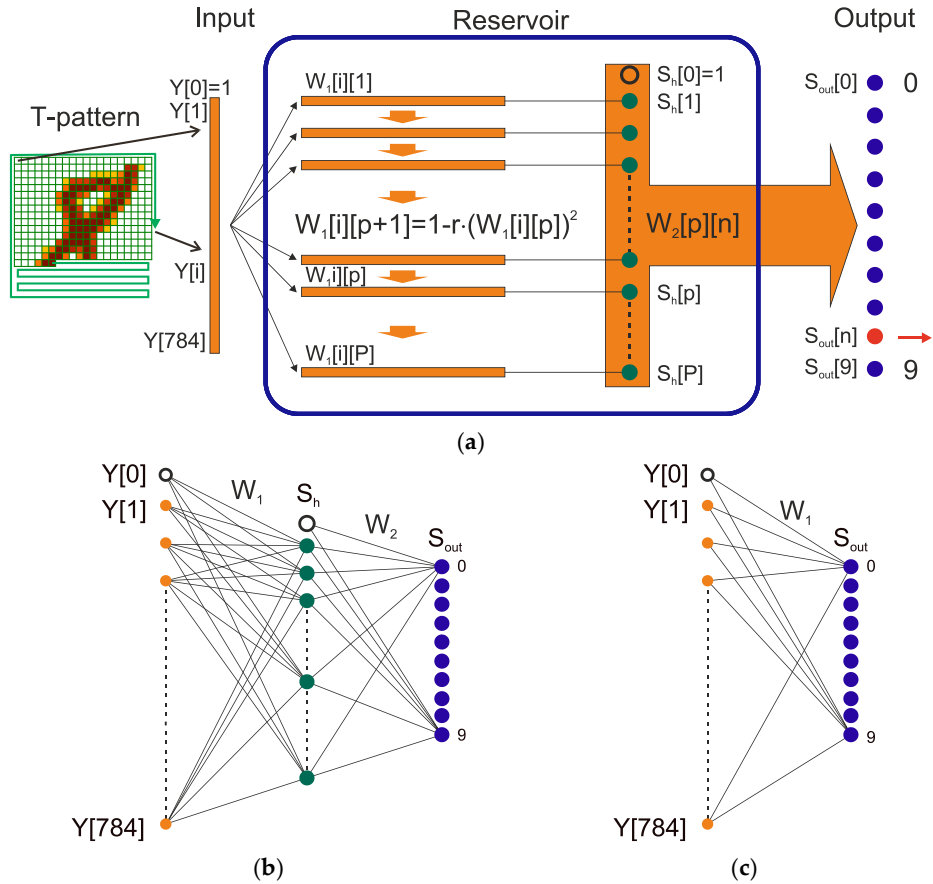

**Figure 2.** LogNNet architecture (**a**). Two-layer (**b**) and 1-layer (**c**) feedforward networks.

The weights $W_1$ in the reservoir were set recursively, based on the logistic mapping

$$x_{p+1} = 1 - r \cdot x_p^2, \tag{2}$$

where *r* is a positive parameter. By applying (2) to the elements of the weight matrix $W_1$, the following equation is obtained:

$$W_1[i][p+1] = 1 - r \cdot (W_1[i][p])^2. \tag{3}$$

The initial values of the elements of the first row $W_1[i][1]$ are set by the equation

$$W_1[i][1] = A \cdot \sin\left(\frac{i}{784} \cdot \frac{\pi}{B}\right), \tag{4}$$

where *A* and *B* are adjustable parameters; *A* = 0.3, *B* = 5.9, *i* = 0–784.

The classification of handwritten digits in the range "0–9" proceeded according to the largest value of the element of the output layer $S_{out}[n]$, where *n* = 0–9. The training of the weights $W_2$ was performed by the error back-propagation method [37] with the learning rate of 0.3. The initial values of the weights $W_2$ were set randomly in the range between −0.5 and 0.5.

Weights $W_1$ are not trained, and only the weights $W_2$ of the output classifier are trained. This approach resembles the operation of a recurrent reservoir network, except the recurrence transformation is applied not to the input data, but to the elements of the matrix $W_1$.

The logistic mapping is widely known as the simplest model that demonstrates the transition to chaotic behavior through a sequence of period doubling bifurcations (Feigenbaum's script [38]). The cascade of period doubling bifurcations in the logistic mapping can be visualized using a phase-parameter diagram, which is called a one-parameter bifurcation diagram (Figure 3).

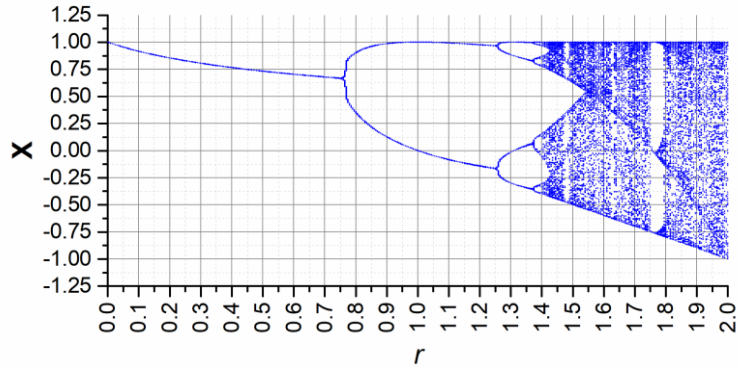

**Figure 3.** The bifurcation diagram of the logistic mapping (2).

In the text, the information on the numbers of network layers and neurons is indicated in the format LogNNet-784:*P*:10, which reflects the numbers of neurons in the input, intermediate and output layers, respectively, without taking the bias neuron into account.

## 3. Results

The dependencies of classification accuracy on the number of training epochs for different T-patterns are shown in Figure 4. With an increase in the number of epochs, the classification accuracy increases and reaches its maximum values when using T-pattern-3. The larger the number of neurons in the hidden layer *P*, the higher the classification accuracy, and for *P* = 100 the accuracy reaches ≈ 89.5%. As T-pattern-3 provided the best results, all subsequent calculations are given using this transformation. The application of transformations to the initial images should be considered as a heuristic technique, which allows one to overcome the disadvantages of the back-propagation of the error learning method. Configuring a neural network is truly considered an art [39], and many authors use affine transformations [40,41] and elastic deformations [42].

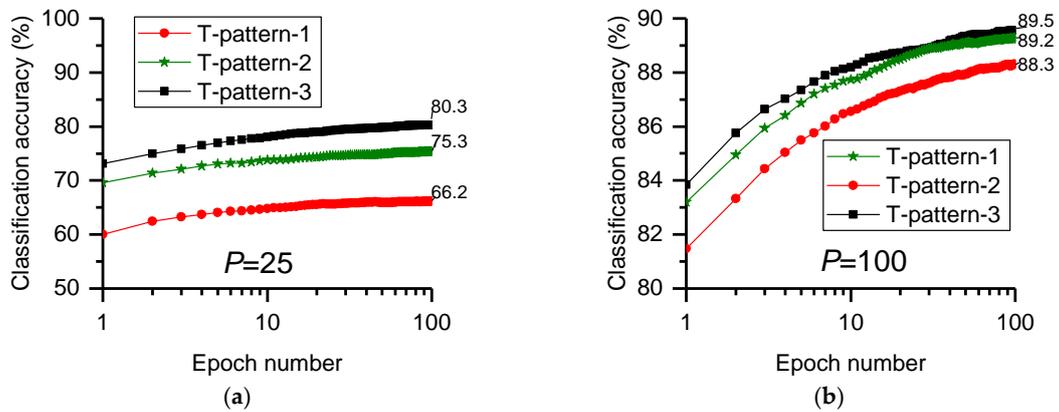

**Figure 4.** The dependence of the classification accuracy on the number of epochs, for different T-patterns. The numbers of reservoir neurons: *P* = 25 (**a**), *P* = 100 (**b**) and *r* = 1.885.

The dependence of the classification accuracy on the parameter *r* is presented in Figure 5. The accuracy grows with increasing *r* and an increasing number of neurons *P*. At values of *r* = 1.65, 1.805, 1.885, 1.967 and 1.992, the local maxima of classification accuracy were observed.

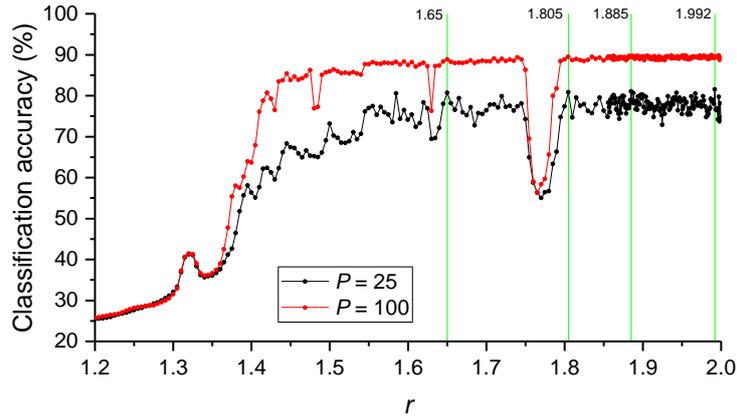

**Figure 5.** The dependence of the classification accuracy on the parameter *r* at *P* = 25 (LogNNet-784:25:10) and *P* = 100 (LogNNet-784:100:10).

The shape of distribution of weights $W_1$ for different reservoir neurons, at *r* = 1.885, is presented in Figure 6 (for convenience of visualization, the inverse transformation T-pattern-1 was applied, and the element $W_1[0][p]$ was placed in the upper left corner). Due to the formation of matrix elements through the logistic relation (2), the kernel's shape changes with each step *p*. Kernels transform the input space to the higher-dimensional feature space. The result of the kernels' operation is released by the reservoir neurons and transmitted to the output classifier.

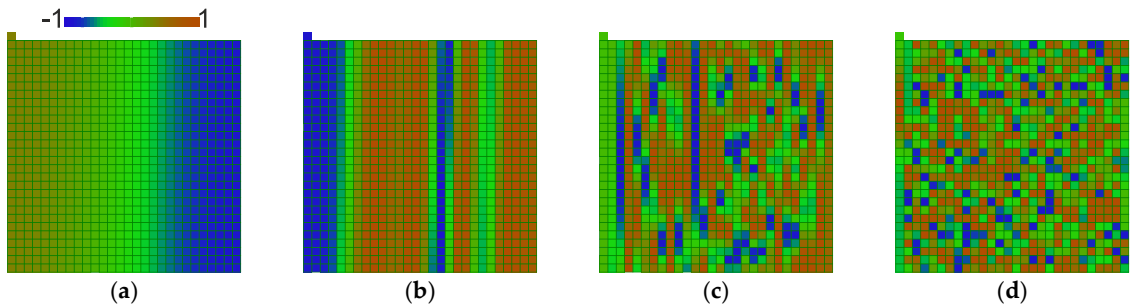

**Figure 6.** The distribution shape of the weights of the matrix $W_1$ for *p* = 6 (**a**), *p* = 10 (**b**), *p* = 15 (**c**), *p* = 25 (**d**) for LogNNet-784:25:10 and *r* = 1.885.

The advantage of LogNNet is that the weights $W_1$ do not have to be stored in a large two-dimensional array with the number of elements $N = 785 \times P$. The weights can be sequentially calculated during the algorithm's operation, at the training and testing stages. For calculation, only three parameters *r*, *A* and *B* should be known (Algorithm 1; see Appendix A.6.1), and only one memory cell $N = 1$ should be reserved with the capacity $MeW_1$ = 4 B (a single type occupies 4 bytes). In practice, it is convenient to reserve a one-dimensional array $W_{1\_alg2}[i]$, where *i* = 0–784, and recalculate its values with a subsequent increase in *p*, $W_{1\_alg2}[i] = 1 - r \cdot (W_{1\_alg2}[i])^2$ (Algorithm 2; see Appendix A.6.2). In this case, the maximum reserved memory is estimated as $MeW_1 \approx 3$ kB. If each element $W_1$ would be stored separately in memory, then, for *P* = 100, *N* = 78500, the array would occupy the memory $MeW_1 \approx 306$ kB (Algorithm 3; see Appendix A.6.3). Therefore, memory savings to use Algorithm 1 and Algorithm 2 are obvious. To evaluate the total memory, it is necessary to sum over all layers of the network $MeW = MeW_1 + MeW_2 + ... + MeW_x$.

Table 1 gives comparative estimates of the classification accuracy and the amount of allocated memory for the weight arrays for various network configurations, including for LogNNet with 1–2 layer output classifiers, and for 1–2 layer feedforward networks (Figure 2b,c). Applying Algorithm 1, *MeW* values can be reduced by 3 kB, and reach $MeW \approx 1$ kB for LogNNet-784:25:10, although Algorithm 1 is slower than Algorithm 2.

**Table 1.** Comparison of LogNNet network with other machine learning approaches. For all LogNNet configurations $r$ = 1.885, $A$ = 0.3, $B$ = 5.9.

| Classifier | MNIST-10 Accuracy | Memory *MeW* |
|---|---|---|
| LogNNet-784:25:10 | 80.3% | (785 + 26·10) ·4 B = 4 kB (Alg. 2) or 1 kB (Alg. 1) |
| LogNNet-784:100:10 | 89.5% | (785 + 101·10) ·4 B = 7 kB (Alg. 2) or 4 kB (Alg. 1) |
| LogNNet-784:200:10 | 91.3% | (785 + 201·10) ·4 B = 10.9 kB (Alg. 2) or 7.8 kB (Alg. 1) |
| LogNNet-784:100:60:10 | 96.3% | 7455·4 B = 29 kB (Alg. 2) or 26 kB (Alg. 1) |
| Lin. 1-Layer 784:10 | 90.6% | 785·10·4 B = 30.7 kB |
| Lin. 2-Layer 784:10:10 | 91.9% | 7960·4 B = 31 kB |
| Lin. 2-Layer 784:30:10 | 95.6% | 23860·4 B = 93.2 kB |
| LeNet-1 [40] | 98.3% | ≈ 2500·4 B = 9.7 kB |
| LeNet-5 [40] | 99.05% | ≈ 60000·4 B = 234 kB |
| ESN [41] | 79.43% | ≈ 41000·4 B = 160 kB |
| Bonsai 16 kB [17] | 90.4% | ≈ 16 kB |
| Bonsai 84 kB [17] | 97.01% | ≈ 84 kB |
| ProtoNN [20] | 93.8% | ≈ 16 kB |
| CNN 2 kB [21] | 99.15% | ≈ 2 kB |
| FastGRNN [18] | 98.2% | ≈ 6 kB |
| Spectral-RNN [22] | 97.7% | ≈ 6 kB |
| NeuralNet Pruning [43] | 81% | ≈ 9 kB |
| GBDT [17] | 97.90% | ≈ 5859 kB |

Table 2 summarizes the processing times ($t_{alg1}$, $t_{alg2}$ and $t_{alg3}$) of a single image by the LogNNet network on the IoT device Raspberry Pi 4 for a different number of neurons in reservoir $P$.

**Table 2.** Comparison of processing times of one image on Raspberry Pi 4 for Algorithm 1, Algorithm 2 and Algorithm 3.

| $P$ | 25 | 45 | 75 | 100 |
|---|---|---|---|---|
| $t_{alg1}$, ms | 5.18 | 13.65 | 33.30 | 56.44 |
| $t_{alg2}$, ms | 0.74 | 1.17 | 1.85 | 2.38 |
| $t_{alg3}$, ms | 0.41 | 0.76 | 1.26 | 1.78 |
| $t_{alg1}/t_{alg3}$ | 12.63 | 17.96 | 26.43 | 31.70 |
| $t_{alg2}/t_{alg3}$ | 1.80 | 1.54 | 1.47 | 1.34 |

## 4. Discussion

Logistic mapping can be presented in several formats by applying variable substitution [44]. The most common formats are as follows:

$$\begin{aligned} x_{p+1} &= a \cdot x_p \cdot (1 - x_p) \\ x_{p+1} &= 1 - r \cdot x_p^2 \\ x_{p+1} &= x_p^2 + c \end{aligned}, \quad (5)$$

where $a$, $r$, $c$ are constants, whose variation range is described in [44].

In the current study, the representation of logistic mapping expressed by Equation (2) is used; however, any format from the set (5) can be applied.

Equation (4) was chosen empirically; it sets the initial conditions for filling the array $W_1$, which defines the set of filters (kernels) presented in Figure 6. The sine function in Equation (4) was prompted by the shape of the distribution of weights in feedforward networks (Figure 2b,c). The selection of the optimal coefficients $A$ and $B$, and the format of the optimal function, can be the subject of a separate study.

Figure 4 demonstrates that after the first training epoch, the classification accuracy approaches its maximum value, and differs from the maximum value by no more than 5%. This indicates good convergence of the backpropagation of errors algorithm used in training. In the case of limited computational resources, the number of epochs during training can be significantly reduced.

The shape of the dependence of classification accuracy on the parameter *r* resembles the dependence of the Lyapunov exponent $\lambda$ on *r* [38], and indicates the importance of the chaotic behavior of the logistic mapping in the recognition process. The Lyapunov exponent $\lambda$ represents an effective way of assessing the "chaoticness," and determines the measure of the divergence of the trajectory—that is, the nature of the trajectory change depending on the change in the initial condition. Positive $\lambda > 0$ corresponds to chaotic behavior with diverging trajectories, while $\lambda < 0$ corresponds to converging trajectories and no chaos. The comparison of the dependence of the Lyapunov exponent for the logistic mapping (2) and the classification accuracy of the LogNNet-784: 100:10 network on the parameter *r* is presented in Figure 7. The displayed dependencies are calculated for the same set of *r* values.

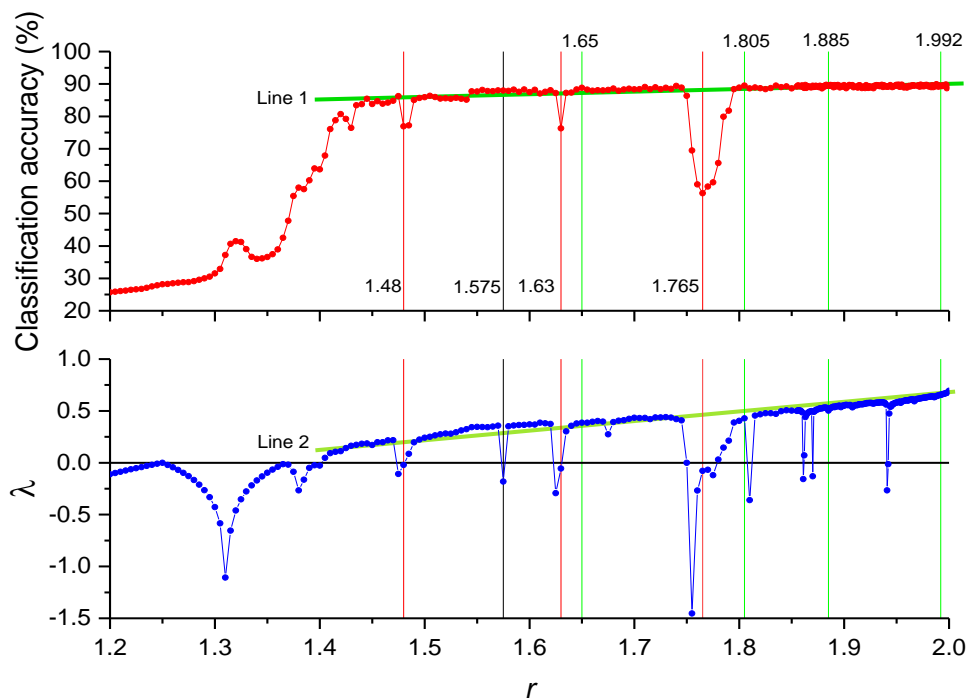

**Figure 7.** Comparison of classification accuracy (LogNNet-784:100:10) and Lyapunov exponent $\lambda$ on the parameter *r*. Vertical green marks with labelled *r* values correspond to local maxima of classification accuracy values, red marks correspond to local minima. One of the local minima of the Lyapunov exponent values is marked with a black mark. Line 1 is an approximating line for classification accuracy values. Line 2 is an approximating line for $\lambda$ values.

An increasing trend of classification accuracy change on *r* (positive slope of Line 1) can be observed, with maximum values in the range of *r* = 1.4–2. This correlates with the positive slope of $\lambda$ change on *r* (Line 2), where positive values of $\lambda$ prevail at *r* > 1.4. Therefore, an increase in the chaoticness of the logistic mapping leads to an increase in the classification accuracy, and the highest values of classification accuracy were observed at *r* > 1.4.

Although a correlation between classification accuracy and the Lyapunov exponent exists, it is not strict. For example, local minima $\lambda$ at *r* = 1.48, 1.63, 1.765 correspond to local minima of classification accuracy (red marks in Figure 7). However, there are a number of minima $\lambda$, for example, for *r* = 1.575, for which the classification accuracy remains high (black mark in Figure 7). The widest minimum $\lambda$ in the vicinity of *r* = 1.765 also corresponds to the widest minimum of the classification accuracy. Local maxima of classification accuracy, for example, for *r* = 1.65, 1.805, 1.885, correlate with even weaker $\lambda(r)$. Nevertheless, in a number of cases, for example, for *r* = 1.65, 1.805, local maxima of classification accuracy arise at the border of strong chaos and the ordered behavior of the logistic mapping. The important role of the chaotic dynamics of the reservoir was highlighted in the studies on ESNs (setting the spectral radius) [29,45], and LSMs (setting separation property) [30]. A detailed study of the influences of the chaos parameters of the logistic mapping on the classification accuracy of the LogNNet neural network can be a topic for future research.

The important role of chaotic dynamics in the presented network architecture calls for a hardware implementation of the network reservoir, using, for example, analog circuits of chaotic oscillators [46,47]. In this way, the IoT device processor is released from complex calculations of reservoir states. This approach is a promising subject for the future studies.

The shape of the filters presented in Figure 6 depends on the parameters of Equations (2) and (4) and on the number of the neurons of the hidden layer $p$. In future studies, the classification properties of the network can be improved, for example, by mixing rows (columns) of filters with different values of $p$, by applying a different parameter $r$ or by applying a combination of logistic mapping equations in different parts of the filter. Such transformations could lead to discovering more suitable filter configurations for each specific classification problem.

The main practical advantage of LogNNet networks is that their implementation algorithm, based on logistic mapping, can significantly save memory resources by transferring the computational load to the processor, which calculates the coefficients of the matrix $W_1$ using Equations (1)–(4). Three algorithms have been developed: Algorithm 1, Algorithm 2 and Algorithm 3. A detailed description of algorithms' pseudocodes is given in the Appendix (see Appendix A.6.1, A.6.2 and A6.3). In all algorithms, the output of the network has the same value. The main difference lies in the amount of memory consumed and in the speed of the algorithms. Algorithm 1 uses only one auxiliary variable $W_{1\_alg1}$ (see A.6.1), Algorithm 2 uses an auxiliary one-dimensional array $W_{1\_alg2}[i]$ (see A.6.2) and Algorithm 3 uses a two-dimensional array $W_1[i,j]$ (see A.6.3). The advantage is especially noticeable when comparing LogNNet networks with feedforward and reservoir networks (Table 1). For example, Lin. 2-Layer 784:10:10 network and LogNNet-784:200:10 network have almost the same classification accuracy of ≈ 91.5%, but the LogNNet-784:200:10 network requires almost three times less *MeW* memory. If Algorithm 1 is used instead of Algorithm 2, then the *MeW* value for LogNNet is reduced by another 3 kB. The amount of memory consumed by LogNNet-784:25:10 can be reduced from 4 kB to 1 kB using Algorithm 1. At the same time, the classification accuracy of ≈ 80% corresponds to the classification accuracy of ESN reservoir networks (see Table 1), but the *MeW* memory used by LogNNet-784:25:10 is almost 160 times smaller. Compared to convolutional (LeNet-1, CNN 2 kB) and recurrent (FastGRNN, Spectral-RNN) networks, the advantage is not so obvious. However, the potential of LogNNet is not fully revealed, and requires a comprehensive study. In addition, convolutional networks require complex program code, while LogNNet is a relatively simple network to implement on embedded systems, such as Arduino or Raspberry Pi and other constrained devices. Although LogNNet provides almost the same values of classification accuracy and memory used with the Bonsai and ProtoNN algorithms, the principle of the algorithms is fundamentally different. Table 1 does not include the characteristics of the BonsaiOpt algorithm [17], which demonstrates outstanding classification accuracy of ≈ 95% at 2 kB RAM, because BonsaiOpt applies strong optimization with the exception of floating point operations using 1 byte number format. Similar optimization applied to the LogNNet algorithm can significantly reduce the amount of memory used by the algorithm and is a subject for the future research. The GBDT algorithm employs significantly more memory than LogNNet-784:100:60:10; however, the classification accuracy of GBDT is insignificantly higher. NeuralNet Pruning provides a lower classification accuracy than LogNNet. Therefore, in comparison with these algorithms, the use of LogNNet is preferable.

The results of performance measurements of the Algorithm 1 ($t_{alg1}$), Algorithm 2 ($t_{alg2}$) and Algorithm 3 ($t_{alg3}$) operation on Raspberry Pi 4 are presented in Table 2. For any $P$, Algorithm 1 requires the largest amount of time to process the image. The ratio $t_{alg1} > t_{alg2} > t_{alg3}$ was observed, and it is obviously explained by a higher speed of reading the weights $W_1$ from RAM memory, relative to the speed of weights recalculation by the processor using the Equations (3) and (4). The $t_{alg1}/t_{alg3}$ ratio increases with increasing *P* because of the growing load on the processor and the slowdown in Algorithm 1 caused by recalculation of weights. Another interesting observation is that with an increase in the number of neurons in the hidden layer *P*, the $t_{alg2}/t_{alg3}$ ratio decreases. A possible reason may be a nonlinear increase in the access time to RAM memory with an increase in the size of data arrays. The processing times of one image by the presented algorithms are in the range of ≈ 0.4–50 ms. It allows MNIST-10 images to be processed on a Raspberry Pi 4 with a frame rate in the kilohertz range. Further technical aspects of the distribution of computing resources in constrained devices for LogNNet implementation can be the subject of future research.

Currently, more than 300 databases [48] are available for training neural networks and classification algorithms in different areas: pattern recognition in images and video data (for example, MNIST [36], CiFAR [49], Chars74 K [50]), sound recognition, weather forecasting and medical and biological research. The most critical tasks to solve on constrained devices in the IoT industry are the following: face recognition [51], speech recognition [52], health monitoring using human activity recognition on mobile devices [53] and autonomous vehicle driving [54]. This study is limited to the testing of LogNNet network on the database of handwritten numbers MNIST. However, the universal feature of the reservoir to transform the input data into a higher dimensionality of space (kernel trick [31]) can be applied to other problems of classification and prediction. The future research may focus on testing other popular databases and expanding the areas of LogNNet use in the IoT industry. Special attention should be given to the study of the transfer learning technique for the LogNNet configuration; when on one database, a trained network can be successfully used on another database, and it can reduce the cost of training on resource-constrained devices [55–57]. For the LogNNet network, it can be implemented as the use of the same parameters of the network reservoir for different databases, or the use of several reservoirs, configured to extract universal features of the input information.

## 5. Conclusions

The study presents the LogNNet neural network, which uses filters based on logistic mapping to achieve high accuracy indicators for the classification of handwritten digits from the MNIST-10 database. The network uses less memory for weight arrays than many well-known network configurations. The future research directions may search for other chaotic mappings and initial conditions, for more efficient operation of LogNNet. From a research perspective, LogNNet is interesting for studying the fundamental issues of the influence of chaos on the behavior of reservoir neural networks. The proposed neural network can be used to implement artificial intelligence based on constrained devices with limited memory size, which are integral blocks for the creation of AmI and modern IoT environments.

**Supplementary Materials:** The following are available online at www.mdpi.com/xxx/s1: an example of executable code of the LogNNet network (LogNNet.zip).
**Author Contributions:** Conceptualization, A.V.; software, A.V.; writing—original draft preparation, A.V.; project administration, A.V.; Author have read and agreed to the published version of the manuscript.
**Funding:** This research was supported by the Russian Science Foundation (grant number 16-19-00135).
**Acknowledgments:** The author expresses his gratitude to Andrei Rikkiev for the valuable comments in the course of the article's translation and revision. Special thanks to the editors of the journal and anonymous reviewers for constructive criticism and improvement suggestions.
**Conflicts of Interest:** The authors declare no conflict of interest.

## Appendix A

The pseudocode demonstrates the training and testing methodology for the LogNNet network based on the MNIST-10 handwritten digits database. The algorithm was implemented in the Delphi programming language, and instructions for running the code are given in Supplementary Materials.

**A1.** Initialization of basic constants, types, arrays, variables and functions.

```
const
    N_test = 10000;//The number of elements in the test data array from the MNIST-10
                database
    N_train = 60,000;//The number of elements in the training data array from the MNIST-
                10
    Y_max = 784;//The number of dots in the image of a handwritten digit
    P_max = 25;//Number of neurons in the hidden layer
    N_max = 9;//Numbering limit of output neurons 0..9
type
    InputData = array [0..Y_max] of byte;//Input data type
    InputData2 = array [0..Y_max] of real;//Input data type after normalization
    OutputData = array [0..N_max] of real;//Output data type
    Hiddenlayer = array [0..P_max] of real;//Data type of the hidden layer
var
    Train_data: array [1..N_train] of InputData;//Training data array from the MNIST-10
                                    database
    Test_data: array [1..N_test] of InputData;//Test data array from the MNIST-10 database
    Test_label: array [1..N_test] of byte;//Array of test data labels
    Y: InputData2;//Array of input image data
    Sh: Hiddenlayer;//Array of hidden layer data
    Sout: OutputData;//Array of output layer data
    W1: array [0..Y_max, 1..P_max] of real;//Array of weights W1
    W1_alg2: array [0..Y_max] of real;//An auxiliary array for calculating the weights W1
                        for Algorithm 2
    W1_alg1: real;//Auxiliary variable for calculating the weights W1 for Algorithm 1
    W2: array[0 .. P_max,0..N_max] of real;//Array of weights W2
    i, j, k, i_max: integer;//Auxiliary variables of integer type.
    Sh_max, Sh_min, Usre: Hiddenlayer;//Auxiliary arrays for normalizing the hidden
                            layer data
    Accuracy: real;//Accuracy of network classification (in percent).
functions
    Function VM_transform (YYY: InputData): InputData;//Function for T-pattern
                                    transformation
    Function Normalization (XX: InputData): InputData2;//Function to normalize input
                                    data
    Function Fout (R: real): real;//Logistic activation function for output neurons
    begin
        Fout: = 1/(1 + exp (-R));
    end;
```

**A2.** Loading data files and setting initial conditions.

Zeroing used arrays.
Loading training *Train_data*, test data *Test_data* (procedure TForm1.Load_mnist_buttonClick in Supplementary Materials).
Loading T-Pattern files (procedure TForm1.Load_Tpattern_buttonClick in Supplementary Materials).
Entering *r*, *A*, *B* values and the number of epochs to train the network.

**A3.** Filling in arrays $W_1$ and $W_2$.

```
//The initial values of the elements of the first row W1[i][1] using Equation (4)
j:=1;
for i:=0 to Y_max do
begin
    W1[i,j]:= A*sin((i/Y_max)*Pi/B);
end;

//Applying Equation (2) to the elements of the weight matrix W1
for j:=2 to P_max do
begin
    for i:=0 to Y_max do
    begin
        W1[i,j]:=(1-r*W1[i,j-1]*W1[i,j-1]);
    end;
end;

//The initial values of the weights W2 are set randomly between -0.5 and 0.5.
for i:=0 to P_max do
    for j:=0 to N_max do W2[i,j]:=(0.5-Random);
```

**A4.** Calculation of auxiliary data for normalizing the values of the hidden layer.

Auxiliary data for calculating $f_h$ and subsequent normalization are arrays of maximum *Sh_max*, minimum *Sh_min* and average values *Usre* of the weighted sum of the hidden layer for all training data. The algorithm for calculating auxiliary data is given in Supplementary Materials (procedure TForm1.Fh_activ_fun_parametersClick).

**A5.** Training the array $W_2$ using the backpropagation of error method.

For training the array *$W_2$* using the backpropagation of error method, see procedure TForm1.EpochTrainingClick and procedure TForm1.BackPropagationClick in Supplementary Materials.

**A6.** Testing the network.

The calculation of the classification accuracy on the MNIST-10 test data can be performed using one of three algorithms: Algorithm 1, Algorithm 2 or Algorithm 3.
```
Correct_test:=0;//Counter of correct classifications of the network using test data
for Nom_tren:=1 to N_test do//Loop over MNIST-10 test data
begin
//Applying T-Pattern transformation and normalizing the input image of the test data
    Y:=Normalization(VM_transform(Test_data[Nom_tren]));
```

**A6.1.** Algorithm 1.

```
for j:=1 to P_max do//Loop over the neurons of the hidden layer
begin
Sh[j]:=0;
for i:=0 to Y_max do
    begin
    W1_alg1:= A*sin((i/Y_max)*Pi/B);
    for k:=2 to j do W1_alg1:=(1-r*W1_alg1*W1_alg1) ;
    Sh[j]:=Sh[j]+Y[i]*W1_alg1;
    end;
Sh[j]:=((Sh[j]-Sh_min[j])/(Sh_max[j]-Sh_min[j])-0.5)-Usre[j];//Normalizing the hidden layer
end;
```

**A6.2.** Algorithm 2.

```
for j:=1 to P_max do//Loop over the neurons of the hidden layer
begin
Sh[j]:=0;
for i:=0 to Y_max do
begin
    if j=1 then W1_alg2[i]:= A*sin((i/Y_max)*Pi/B)
        else W1_alg2[i]:=(1-r*W1_alg2[i]*W1_alg2[i]) ;
    Sh[j]:=Sh[j]+Y[i]*W1_alg2[i];
end;
Sh[j]:=((Sh[j]-Sh_min[j])/(Sh_max[j]-Sh_min[j])-0.5)-Usre[j];//Normalizing the hidden layer
end;
end;
```

**A6.3.** Algorithm 3.

```
for j:=1 to P_max do//Loop over the neurons of the hidden layer
begin
    Sh[j]:=0;
    for i:=0 to Y_max do Sh[j]:=Sh[j]+Y[i]*W1[i,j];//W1 values are pre-filled, see A3.
    Sh[j]:=((Sh[j]-Sh_min[j])/(Sh_max[j]-Sh_min[j])-0.5)-Usre[j];//Normalizing the hidden
layer end;
```

**A7.** Calculation of network output and classification accuracy.

```
Sh[0]:=1;//Determining the value of the bias neuron of the hidden layer
for j:=0 to N_max do//Loop over the neurons of the output layer
begin
    Sout[j]:=0;
    for i:=0 to P_max do Sout[j]:=Sout[j]+Sh[i]*W2[i,j];
```

```
            Sout[j]:=Fout(Sout[j]);
        end;
            i_max:=1;//Index of the output neuron with the maximum value
            For i:=0 to N_max do if Sout[i]>Sout[i_max] then i_max:=i;
            //Counting the number of correct recognitions
            if i_max=Test_label[Nom_tren] then Correct_test:=Correct_test+1;
        end;//End of the cycle for test data MNIST-10
            Accuracy:=(Correct_test/N_test)*100;//Calculation of classification accuracy
```